\documentclass[conference]{IEEEtran}
\IEEEoverridecommandlockouts
\usepackage{cite}
\usepackage{amsmath,amssymb,amsfonts}
\usepackage{algorithmic}
\usepackage{graphicx}
\usepackage{textcomp}
\usepackage{xcolor}
\usepackage{booktabs}
\usepackage{hyperref}
\usepackage[none]{hyphenat}

\def\BibTeX{{\rm B\kern-.05em{\sc i\kern-.025em b}\kern-.08em sT\kern-.1667em\lower.7ex\hbox{E}\kern-.125emX}}

\begin{document}

\title{Downscaling Precipitation with Bias-informed Conditional Diffusion Model}

\author{\IEEEauthorblockN{Ran Lyu}
\IEEEauthorblockA{
\textit{Virginia Tech}\\
ran22@vt.edu}
\and
\IEEEauthorblockN{Linhan Wang}
\IEEEauthorblockA{
\textit{Virginia Tech}\\
linhan@vt.edu}
\and
\IEEEauthorblockN{Yanshen Sun}
\IEEEauthorblockA{
\textit{Virginia Tech}\\
yansh93@vt.edu}
\and
\IEEEauthorblockN{Hedanqiu Bai}
\IEEEauthorblockA{
\textit{Texas A\&M University}\\
baisy@tamu.edu}
\and
\IEEEauthorblockN{Chang-Tien Lu}
\IEEEauthorblockA{
\textit{Virginia Tech}\\
clu@vt.edu}
}

\maketitle

\begin{abstract}
Climate change is intensifying rainfall extremes, making high-resolution precipitation projections crucial for society to better prepare for impacts such as flooding. However, current Global Climate Models (GCMs) operate at spatial resolutions too coarse for localized analyses. To address this limitation, deep learning-based statistical downscaling methods offer promising solutions, providing high-resolution precipitation projections with a moderate computational cost. In this work, we introduce a bias-informed conditional diffusion model for statistical downscaling of precipitation. Specifically, our model leverages a conditional diffusion approach to learn distribution priors from large-scale, high-resolution precipitation datasets. The long-tail distribution of precipitation poses a unique challenge for training diffusion models; to address this, we apply gamma correction during preprocessing. Additionally, to correct biases in the downscaled results, we employ a guided-sampling strategy to enhance bias correction. Our experiments demonstrate that the proposed model achieves highly accurate results in an \(8 \times\) downscaling setting, outperforming previous deterministic methods. The code and dataset are available at \href{https://github.com/RoseLV/research_super-resolution}{Github}.
\end{abstract}

% \vspace{-5pt}
\section{Introduction}
% \vspace{-5pt}

Due to climate change, there is a growing demand for reliable weather and climate simulation at local scales \cite{kumar2023modern}. The current de facto technique Global Climate Models (GCMs) \cite{heavens2013studying} can simulate the Earth's response to varying atmospheric greenhouse gas (GHG) emissions scenarios. However, the outputs from GCMs are often coarse and lack the granularity \cite{gutowski2020ongoing} needed to understand local weather patterns and their impacts on specific sectors such as agriculture, water resources, and food security.

One of the foundational methods is statistical downscaling, which enhances coarse climate model outputs by predicting fine-resolution data based on statistical relationships between low-resolution and high-resolution observations. This approach shares similarities with image super-resolution techniques in computer vision. A historically significant image super-resolution method is the Super-Resolution Convolutional Neural Network (SRCNN). Although SRCNN has been widely adopted and dominated the field for nearly a decade, it suffers from limitations such as a restricted receptive field and insufficient network depth, which impede its ability to capture global context and effectively extract features \cite{luo2016understanding}. Additionally, SRCNN struggles with poor adaptation to long-tail data distributions \cite{feldman2020neural}, such as precipitation, making it unsuitable for tasks like 8× downscaling.

Recently, denoised diffusion models \cite{ho2020denoising, song2020score} have emerged as the dominant deep generative models for images due to their comprehensive coverage of data distributions and high-quality outputs. Improved Denoising Diffusion Probabilistic Models (DDPMs)\cite{nichol2021improved} are generative models that produce high-quality samples, achieve competitive log-likelihoods with simple modifications, enable faster sampling through learned variances, and scale effectively with model capacity and compute. Furthermore, diffusion models outperform state-of-the-art generative models like GANs\cite{goodfellow2020generative} in image synthesis tasks by leveraging improved architectures and classifier guidance for enhanced fidelity and diversity, achieving superior FID scores across multiple resolutions while maintaining efficient sampling and better distribution coverage\cite{dhariwal2021diffusion}. 

In this work, we aim to adapt conditional diffusion models for the task of precipitation downscaling. However, precipitation downscaling has its unique challenges. \textbf{1)} The distribution of precipitation is inherently non-normal, posing difficulties for traditional deep learning models to capture its long-tail characteristics. \textbf{2)} Bias is a critical metric for precipitation downscaling, as small inaccuracies can significantly impact downstream applications. To tackle these issues, we introduce a Bias-aware Guided Sampling (BGS) approach to systematically reduce bias during the downscaling process, improving the overall accuracy and reliability of the generated high-resolution precipitation data.

% \vspace{-5pt}
\section{Method}
% \vspace{-5pt}

In this section, we first introduce the definition and statistical downscaling. Then we explain how to train a conditioned denoised diffusion model for statistical downscaling. Finally, we introduce the motivation and formulation of our two core innovations: Gamma Correction and Bias-aware Guided Sampling.

\begin{figure*}[htbp]
\centerline{\includegraphics[width=\textwidth]{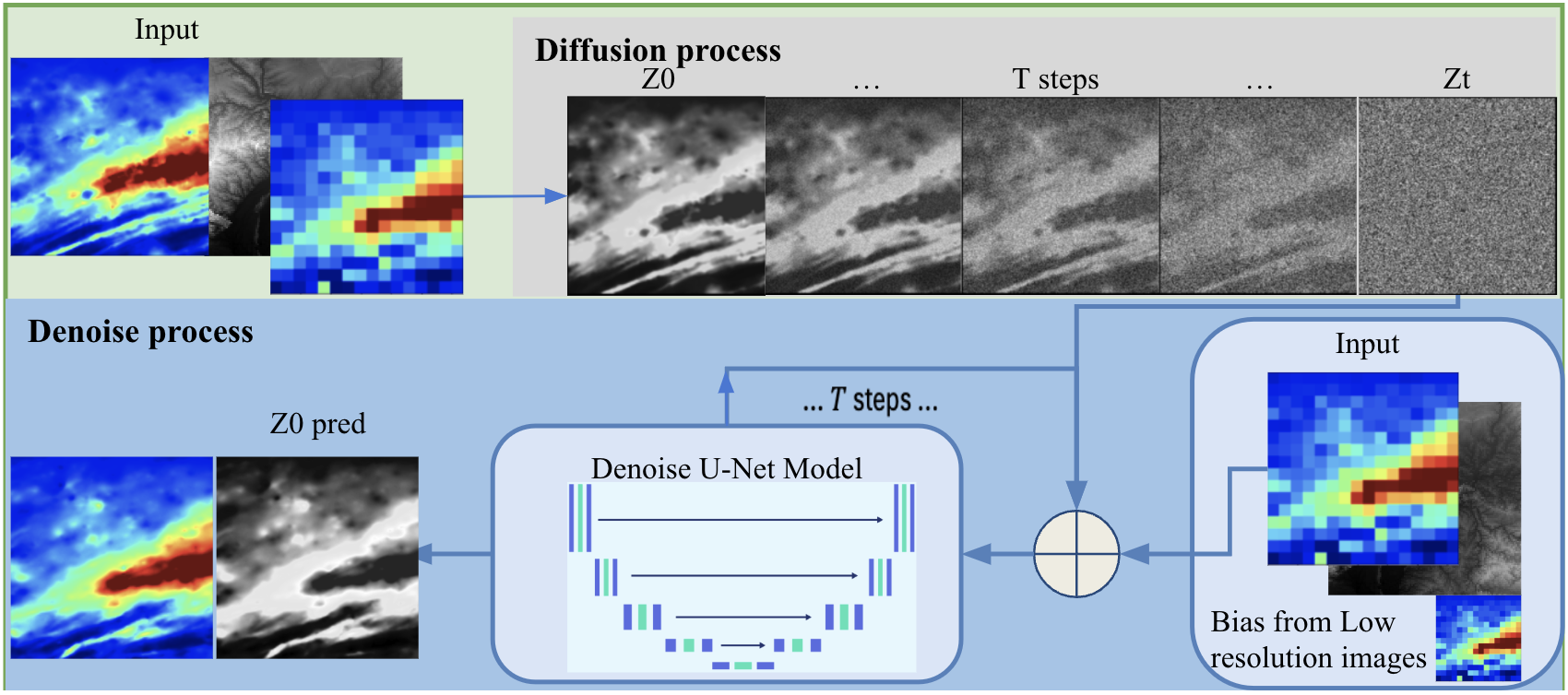}}
\caption{Overall pipeline of the proposed framework: In the training phase, a noise-corrupted high-resolution image, a low-resolution image, and a topography image are concatenated and input into a U-Net model. The U-Net is trained to predict the noise in the corrupted high-resolution precipitation image. In the prediction phase, the trained U-Net iteratively denoises precipitation images, transforming pure noise into high-resolution precipitation. Our bias-informed sampling strategy quantifies and reduces the bias between the corrupted high-resolution precipitation and the low-resolution input at each denoising step.}
\label{fig:fullwidth}
\end{figure*}

\subsection{Problem Setting and Conditional Diffusion Models}

Statistical downscaling involves generating high-resolution climate variables from low-resolution counterparts using statistical methods. This process closely resembles the image super-resolution task in computer vision.\cite{saharia2022image}.

We are given a dataset of low-high resolution precipitation pairs, denoted \(\mathcal{D} = \{\boldsymbol{x}_i, \boldsymbol{y}_i\}^N_{i=1}\), which represent samples drawn from an unknown conditional distribution \(p(\boldsymbol{y}|\boldsymbol{x})\). Here \(\boldsymbol{y}\) represents high-resolution precipitation and \(\boldsymbol{x}\) represents low-resolution precipitation. Due to the high correlation between topography information and precipitation\cite{daly1994statistical}, we concatenate it with \(\boldsymbol{x}\) as the input for our model.

The conditional diffusion model generates a high resolution precipitation \(\boldsymbol{y}_0\) in \(T\) refinement steps. Starting with an initial noise precipitation \(\boldsymbol{y}_T \sim \mathcal{N}(\boldsymbol{0}, \boldsymbol{I})\), the model progressively refines the precipitation through successive iterations \((\boldsymbol{y}_{T-1}, \boldsymbol{y}_{T-2},...,\boldsymbol{y}_0)\) using learned conditional transition distributions \(p_{\theta}(\boldsymbol{y}_{t-1}|\boldsymbol{y}_t, \boldsymbol{x})\) such that \(\boldsymbol{y}_0 \sim p(\boldsymbol{y}|\boldsymbol{x})\).

\subsection{Gamma correction}

Distribution of precipitation naturally deviates from Gaussian distribution, which is favored by deep learning models. Inspired by the similarity between distributions of precipitation and low-light images \cite{li2023pixel}, we apply Gamma Correction on both low and high resolution precipitation data in preprocessing stage. The equation of Gamma Correction is
\begin{equation}
\hat{a} = a^\gamma, \gamma = 0.15 \label{eq}
\end{equation}

This approach addresses the issue of large precipitation areas dominating the optimization process, thereby effectively enhancing the model's performance.

\subsection{Bias-aware Guided Sampling(BGS)}

Classifier-guided sampling is a technique used in \cite{dhariwal2021diffusion}. This approach can improve the performance of diffusion model without slowing down the training process. We propose to incorporate this technique in our sampling process.

While the diffusion process conditions on low-resolution image \(\boldsymbol{x}\), we assume that the information of \(\boldsymbol{x}\) is not fully leveraged due to indirect connection between \(\boldsymbol{x}\) and \(\boldsymbol{y}\). Moreover, the \(\boldsymbol{x}\) comes from GCMs, which are expert models with minimum bias from real climate variables. To fully leverage this prior knowledge, we propose a Bias-aware Guided Sampling approach\cite{dhariwal2021diffusion}. Specifically, we use a L2 norm \(f = \|\boldsymbol{y}_t - \boldsymbol{x}\|_2\) to quantify the bias between \(\boldsymbol{y}_t\) and \(\boldsymbol{x}\) and use its gradients to steer the generation process at each diffusion step. Formally,
\begin{equation}
\boldsymbol{y}_{t-1} = \frac{1}{\sqrt{\alpha_t}}(\boldsymbol{y}_t - \frac{1-\alpha_t}{\sqrt{1 - \Bar{\alpha}_t}}\epsilon_{\theta}(\boldsymbol{y}_t, \boldsymbol{x}, t)) - w \nabla f
\end{equation}

where \(w = 100\) is a hyperparameter we searched by experiments.

\section{Experiment and results}

% \vspace{-5pt}

\subsection{Dataset}

The high-resolution precipitation data(4km) is from PRISM Climate data\cite{daly2008physiographically}. The low-resolution precipitation data(32km) is bilinearly downsampled 8x from the high-resolution data. We use the data between 2000-2018 for training and 2019-2022 for evaluation.

% \vspace{-10pt}

\begin{table}[h]
\centering
\caption{Evaluation of downscaling methods on PRISM dataset}
\label{tab:table}
\vspace{-10pt}
\begin{tabular}{l|ccc}
\toprule
Methods           & RMSE(mm/day)  & Corr   & Bias(mm/day)    \\ \midrule
Interpolation     & 3.134 & 0.939  & \textbf{-0.0053}   \\ 
SRCNN             & 5.921 & 0.828  & 0.0660  \\ 
Ours w/o BGS, Topo & 3.019 & 3.943  & 0.0544  \\ 
Ours w/o BGS       & \underline{2.991} & \underline{0.944}  & -0.0548 \\ 
Ours              & \textbf{2.972} & \textbf{0.945} & \underline{-0.0389} \\ \bottomrule
\end{tabular}
\end{table}

% \vspace{-10pt}

\begin{figure}[h]
\centerline{\includegraphics[width=0.5\textwidth]{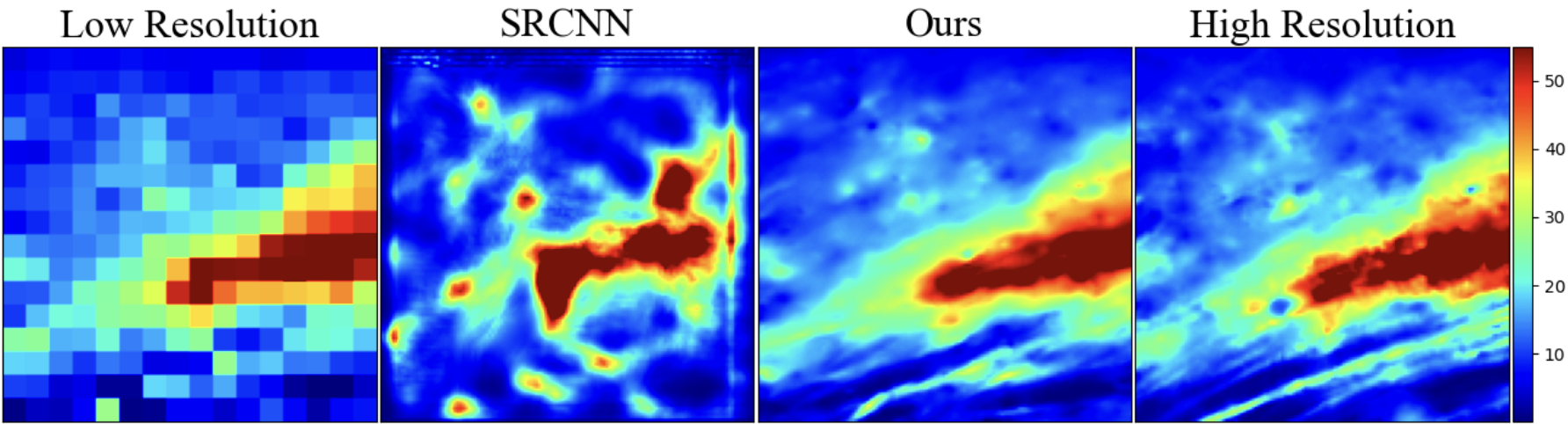}}
\vspace{-10pt}
\caption{One example of \(16 \times 16 \rightarrow 128 \times 128\) downscaling of precipitation.}
\label{fig:figure}
\end{figure}

% \vspace{-10pt}

\subsection{Evaluation metrics}

Following DeepSD \cite{vandal2017deepsd}, we evaluate our method and baselines using three metrics: RMSE, correlation, and bias. RMSE measures the average error, correlation assesses the linear relationship and bias quantifies systematic errors. Both RMSE and bias are in mm/day. The interpolation method uses bilinear interpolation to upscale the original low-resolution image to high resolution, serving as our baseline. SRCNN performs poorly in the evaluation metrics, with a significantly increased RMSE and noticeably low correlation, highlighting its limitations in effectively downscaling precipitation data.

\subsection{Implementation details}
one A100 aOur code is based on improved diffusion \cite{nichol2021improved}. We treat both the high and low resolution precipitations as \(128 \times 128\) grayscale images. All the precipitations are cropped from the same region for convenience. We use AdamW \cite{loshchilov2017decoupled} and learning rate 3e-4. We train our model using one A100 for 1000 epochs, which takes around one day.

\subsection{Quantitative and qualitative results}

The quantitative results are presented in Tab \ref{tab:table}. As shown, our method outperforms the deterministic baseline SRCNN\cite{dong2014learning, vandal2017deepsd}, and both the incorporation of topography information (Topo) and Bias-aware Guided Sampling (BGS) prove effective in this setting.

As shown in Fig \ref{fig:figure}, while SRCNN struggles to generate high-resolution precipitation under an \(8 \times\) super-resolution setting. The distribution and visual performance of the SRCNN image significantly deviate from the high-resolution ground truth, while our method generates physically and visually plausible results with fine details.

\section{Conclusion}

In this work, we proposed a bias-informed conditional diffusion model for precipitation downscaling, addressing the challenges posed by the long-tail distribution of precipitation and biases in low-resolution input data. By incorporating gamma correction during preprocess and introducing Bias-aware Guided Sampling (BGS), our method achieved significant improvements in accuracy and bias correction compared to baseline methods. The experimental results demonstrated the effectiveness of our approach in producing high-resolution precipitation maps with fine details, making it a promising solution for localized climate impact studies. Future work will explore extending this framework to other climate variables and integrating physical constraints for enhanced generalizability.

% \begin{thebibliography}{00}

\bibliographystyle{plain} % Or another style, depending on your requirement
% \nocite{*} % This will include all entries from references.bib
\bibliography{references} % It assumes your BibTeX file is named references.bib

\begin{thebibliography}{10}

\bibitem{daly2008physiographically}
Christopher Daly, Michael Halbleib, Joseph~I Smith, Wayne~P Gibson, Matthew~K Doggett, George~H Taylor, Jan Curtis, and Phillip~P Pasteris.
\newblock Physiographically sensitive mapping of climatological temperature and precipitation across the conterminous united states.
\newblock {\em International Journal of Climatology: a Journal of the Royal Meteorological Society}, 28(15):2031--2064, 2008.

\bibitem{daly1994statistical}
Christopher Daly, Ronald~P Neilson, and Donald~L Phillips.
\newblock A statistical-topographic model for mapping climatological precipitation over mountainous terrain.
\newblock {\em Journal of Applied Meteorology and Climatology}, 33(2):140--158, 1994.

\bibitem{dhariwal2021diffusion}
Prafulla Dhariwal and Alexander Nichol.
\newblock Diffusion models beat gans on image synthesis.
\newblock {\em Advances in neural information processing systems}, 34:8780--8794, 2021.

\bibitem{dong2014learning}
Chao Dong, Chen~Change Loy, Kaiming He, and Xiaoou Tang.
\newblock Learning a deep convolutional network for image super-resolution.
\newblock In {\em Computer Vision--ECCV 2014: 13th European Conference, Zurich, Switzerland, September 6-12, 2014, Proceedings, Part IV 13}, pages 184--199. Springer, 2014.

\bibitem{feldman2020neural}
Vitaly Feldman and Chiyuan Zhang.
\newblock What neural networks memorize and why: Discovering the long tail via influence estimation.
\newblock {\em Advances in Neural Information Processing Systems}, 33:2881--2891, 2020.

\bibitem{goodfellow2020generative}
Ian Goodfellow, Jean Pouget-Abadie, Mehdi Mirza, Bing Xu, David Warde-Farley, Sherjil Ozair, Aaron Courville, and Yoshua Bengio.
\newblock Generative adversarial networks.
\newblock {\em Communications of the ACM}, 63(11):139--144, 2020.

\bibitem{gutowski2020ongoing}
William~J Gutowski, Paul~Aaron Ullrich, Alex Hall, L~Ruby Leung, Travis~Allen O’Brien, Christina~M Patricola, RW~Arritt, MS~Bukovsky, Katherine~V Calvin, Zhe Feng, et~al.
\newblock The ongoing need for high-resolution regional climate models: Process understanding and stakeholder information.
\newblock {\em Bulletin of the American Meteorological Society}, 101(5):E664--E683, 2020.

\bibitem{heavens2013studying}
Nicholas~G Heavens, Daniel~S Ward, and MM~Natalie.
\newblock Studying and projecting climate change with earth system models.
\newblock {\em Nature Education Knowledge}, 4(5):4, 2013.

\bibitem{ho2020denoising}
Jonathan Ho, Ajay Jain, and Pieter Abbeel.
\newblock Denoising diffusion probabilistic models.
\newblock {\em Advances in neural information processing systems}, 33:6840--6851, 2020.

\bibitem{kumar2023modern}
Bipin Kumar, Kaustubh Atey, Bhupendra~Bahadur Singh, Rajib Chattopadhyay, Nachiketa Acharya, Manmeet Singh, Ravi~S Nanjundiah, and Suryachandra~A Rao.
\newblock On the modern deep learning approaches for precipitation downscaling.
\newblock {\em Earth Science Informatics}, 16(2):1459--1472, 2023.

\bibitem{li2023pixel}
Xiangsheng Li, Manlu Liu, and Qiang Ling.
\newblock Pixel-wise gamma correction mapping for low-light image enhancement.
\newblock {\em IEEE Transactions on Circuits and Systems for Video Technology}, 34(2):681--694, 2023.

\bibitem{loshchilov2017decoupled}
I~Loshchilov.
\newblock Decoupled weight decay regularization.
\newblock {\em arXiv preprint arXiv:1711.05101}, 2017.

\bibitem{luo2016understanding}
Wenjie Luo, Yujia Li, Raquel Urtasun, and Richard Zemel.
\newblock Understanding the effective receptive field in deep convolutional neural networks.
\newblock {\em Advances in neural information processing systems}, 29, 2016.

\bibitem{nichol2021improved}
Alexander~Quinn Nichol and Prafulla Dhariwal.
\newblock Improved denoising diffusion probabilistic models.
\newblock In {\em International conference on machine learning}, pages 8162--8171. PMLR, 2021.

\bibitem{saharia2022image}
Chitwan Saharia, Jonathan Ho, William Chan, Tim Salimans, David~J Fleet, and Mohammad Norouzi.
\newblock Image super-resolution via iterative refinement.
\newblock {\em IEEE transactions on pattern analysis and machine intelligence}, 45(4):4713--4726, 2022.

\bibitem{song2020score}
Yang Song, Jascha Sohl-Dickstein, Diederik~P Kingma, Abhishek Kumar, Stefano Ermon, and Ben Poole.
\newblock Score-based generative modeling through stochastic differential equations.
\newblock {\em arXiv preprint arXiv:2011.13456}, 2020.

\bibitem{vandal2017deepsd}
Thomas Vandal, Evan Kodra, Sangram Ganguly, Andrew Michaelis, Ramakrishna Nemani, and Auroop~R Ganguly.
\newblock Deepsd: Generating high resolution climate change projections through single image super-resolution.
\newblock In {\em Proceedings of the 23rd acm sigkdd international conference on knowledge discovery and data mining}, pages 1663--1672, 2017.

\end{thebibliography}

% \end{thebibliography}
\vspace{12pt}

\end{document}